\documentclass{article}

\usepackage{array}
\usepackage{booktabs}
\usepackage{graphicx} % Required for inserting images
\usepackage[utf8]{inputenc}
\usepackage[a4paper, left=2.5cm, right=2.5cm, top=2.5cm, bottom=2.5cm]{geometry}   
\usepackage{amssymb}
\usepackage{amsmath}
\usepackage{algorithm}
\usepackage{algpseudocode}
\usepackage{enumitem}
\usepackage{multirow}
\usepackage{subcaption}
\usepackage[T1]{fontenc}
\usepackage[numbers]{natbib}
\usepackage{xcolor}
\usepackage{url}
\usepackage{pdflscape}
\usepackage{hyperref}
\usepackage{makecell}
\usepackage{authblk}
\usepackage[textsize=scriptsize,colorinlistoftodos]{todonotes}   % per mettere le note a margine (comando \todo{})
\reversemarginpar
	
\usepackage{bm}

\newcommand{\vect}[1]{\bm{#1}} % vector
\newcommand{\poft}[1]{\times 10^{#1}} % power of ten
\newcommand{\icol}[1]{% inline column vector
  \begin{bmatrix}#1\end{bmatrix}^\intercal%
}

\newcommand{\irow}[1]{% inline column vector
  \begin{bmatrix}#1\end{bmatrix}%
}

\newcommand{\nR}[1]{\mathbb{R}^{#1}}		% real number
\title{Locomotion analysis of a quadruped interacting with the lunar granular surface}
\author{Yash Vyas}
\affil[]{Department of Industrial Engineering, University of Padua}
\date{\today}

\begin{document}

\maketitle

\section*{Abstract}
Deploying legged robots in extra-terrestrial environments includes many challenges due to complex terrain interactions, energy, and thermal constraints. For effective mechanical design of a lunar exploration quadrupedal robot, careful consideration of motor torques, energy expenditure, and cost of transport is required. The lunar surface is composed of granular regolith, which impacts the locomotion of legged robots and their performance. Locomotion algorithms trained with rigid contact assumptions are also ineffective when applied to environments with soft contacts, such as granular surfaces, which can result in instability and poor tracking. In this report, the physical modelling of the granular lunar surface-robot foot contacts is applied to a simulation environment with locomotion trained using Reinforcement Learning. A comparison is conducted between the policy trained on rigid contact and soft contact environments, analysing the gait and locomotion performance metrics. The analysis demonstrates that soft contacts simulating regolith surfaces pose additional challenges for Reinforcement Learning based training, result in a qualitatively different gait, and increase the overall energy expenditure. 

\section{Introduction}
Legged robots are becoming increasingly utilized in many inspection and exploration tasks due to their ability to traverse difficult terrains. The robustness of these robots in unstructured environments has improved with modern locomotion control methods such as model-free Reinforcement Learning (RL) \cite{LeeEtAl2020Learningquadrupedallocomotion,MikiEtAl2022Learningrobustperceptive}. Due to their success in terrestrial environments, legged robots have been proposed for missions for space exploration \cite{KolvenbachEtAl2018ScalabilityAnalysisLegged}. To date, lunar and Mars exploration robots have been wheeled or tracked rovers. However, for the exploration of difficult to reach features such as lunar pits, legged robots are more suitable, as they can traverse both flat and rocky terrain. Despite numerous experiments in simulated extra-terrestrial environments on Earth \cite{ArmEtAl2023Scientificexplorationchallenging,MorrellEtAl2024RoboticexplorationMartian}, legged robots are yet to be deployed for space exploration.

Operating robots in extra-terrestrial bodies poses unique challenges compared to operations on Earth \cite{KolvenbachEtAl2024LeggedSystemsExploration}. The components have to be space-grade to resist much higher levels of radiation due to the lack of a protective magnetic field. A detailed understanding of the terrain material based on data from our existing exploration robots is limited. Phenomena such as magnetically charged dust can enter critical components from kick-up or dust storms (in the case of Mars) and impede their operation. A weaker atmosphere or lack of atmosphere complicates thermal management, as heat cannot be easily dissipated. Furthermore, the energy budget of the exploration robot is constrained by what can be obtained from solar panels. These problems are not unique to legged robots; most previous rover-based missions have also grappled with these challenges.

Missions such as LunarLeaper \cite{KolvenbachEtAl2026LunarLeaper} aim to deploy a 10-15 kg legged robot for exploration of the Marius Hills Pit region on the Moon. This feature is of interest as it is located in an area of extensive past volcanic activity, and observations have suggested it could be the entry to a lava tube. A surface mission is now required to confirm this hypothesis and further investigate its characteristics, as lava tubes could be suitable for building human habitats as part of a long-term human presence on the moon \cite{KalitaEtAl2021EvaluationLunarPits}. 

Numerous mission concept proposals have been made to explore lunar pits using large, costly rovers and cranes; however, these methods would not be able to easily traverse the complex terrain consisting of both dusty slopes and boulders, nor would they be able to lower observation equipment into the pit while maintaining platform stability. In contrast, a small and lightweight legged robot is better suited for traversing the diverse terrain types, safely approaching and deploying scientific instruments inside the pit. 

For such missions, a legged robot platform with an optimal locomotion gait that meets the energy and thermal budget in lunar conditions while mitigating dust intrusion into the scientific instruments and solar panels is required. Furthermore, the lunar surface is composed of regolith, a sand-like, viscous granular substrate that risks destabilizing the legged robot during locomotion due to the flow of the soil itself below the feet \cite{Bartsch2014DevelopmentControlEmpirical,KolvenbachEtAl2022TraversingSteepGranular}.

It is essential that the leg morphology and locomotion controller of this platform facilitate safe and robust traversal through this terrain. Such an analysis is imperative, as the variation between rigid vs. soft contact models for terrains can possibly result in different optimal gaits for different morphologies under RL training. It can also affect which set of morphology parameters results in optimal performance according to the mission requirements. 

In the last decade, many different approaches have been developed for robust and agile control of legged robots operating in complex environments. Earlier methods focused on linear or quadratic optimization-based trajectory planners that explicitly estimated the foot contact forces \cite{WinklerEtAl2018GaitTrajectoryOptimization}. In recent years, Reinforcement Learning (RL) has emerged as the method of choice due to its ability to generalize over different environmental conditions such as contact frictions \cite{MikiEtAl2022Learningrobustperceptive}, as well as its robustness to external disturbances and observation errors \cite{HwangboEtAl2019Learningagiledynamic}. RL policy training for extra-terrestrial gravity environments was validated for the Magnecko robot \cite{LeuthardEtAl2024MagneckoDesignandControl} with a gravity-offload system to simulate moon gravity by Arm et. al. \cite{ArmEtAl2025EfficientLearningBased}. 

This report presents a terradynamic analysis of the locomotion performance of a legged robot traversing lunar terrain, extending on \cite{ArmEtAl2025EfficientLearningBased} through the incorporation of granular surface contacts. By implementing the granular surface and robot foot model in \cite{LiEtAl2013TerradynamicsLeggedLocomotion} into a RL simulation environment, I train and compare locomotion policies for both standard rigid contacts and granular surface contacts. By generating simulation rollouts of lunar surface traversals under the trained RL locomotion policy in both environments, the gait characteristics, as well as the locomotion performance in terms of torques, power consumption, and energy expenditure, are analysed.

The outline of this paper is as follows: section \ref{sec:background} provides background on legged robots for extraterrestrial exploration, Reinforcement Learning (RL) for legged robot locomotion, and terradynamic modelling of granular surface contacts. Section \ref{sec:methods} outlines the method and implementation of the chosen granular surface terradynamic model in the simulation environment, as well as the key features of RL policy training. Section \ref{sec:results} compares and analyses the results for both rigid and granular surface contact environments. Finally, section \ref{sec:conclusion} summarizes the findings from this analysis.  
\section{Background}
\label{sec:background}
\subsection{Lunar exploration using legged robots}
Numerous legged robot lunar explorer concepts have been developed and prototyped, utilizing 3, 4, or even 6 legs. Lauron-V \cite{RoennauEtAl2014LAURONVversatile} and SpaceClimber \cite{Bartsch2014DevelopmentControlEmpirical} are both 4-DOF leg hexapods that offer high levels of locomotion stability through a static gait. In contrast, SpaceBok \cite{ArmEtAl2019SpaceBokDynamicLegged} is a 2-DOF parallel mechanism leg quadruped designed for dynamic gaits. SpaceHopper \cite{TrentiniEtAl2023ConceptStudySmall} facilitates highly energy-efficient jumping gaits through geared high-torque 2-DOF legs in a tetrapod configuration. Locomotion studies on Earth have demonstrated that dynamic gaits reduce energy consumption compared to static gaits. In lower gravity environments such as the Moon, this difference is more pronounced, and pronking or jumping gaits are more feasible for energy-efficient locomotion \cite{KolvenbachEtAl2018EfficientGaitSelection,KolvenbachEtAl2019TowardsJumpingLocomotion}, particularly on steep and granular terrain \cite{KolvenbachEtAl2022TraversingSteepGranular}. 

Within quadruped designs, there are two types of configurations for the hip joints: Mammalian (pitch) and Arachnid (yaw). Although Mammalian configurations are more popular among terrestrial quadrupeds for their faster and more energy-efficient gaits \cite{ChaiEtAl2022surveydevelopmentquadruped}, an Arachnid configuration offers more stability for omnidirectional motion due to its larger support polygon and near radial symmetry \cite{KashiriEtAl2016EvaluationHipKinematics}. For this reason, the Arachnid configuration concept based on the Magnecko robot \cite{LeuthardEtAl2024MagneckoDesignandControl}, as experimented with in \cite{ArmEtAl2025EfficientLearningBased}, is modelled in this analysis.

Lunar exploration missions will follow repeated cycles of locomotion, scientific measurement, and rest/recharge phases \cite{KolvenbachEtAl2026LunarLeaper}. The locomotion phase constitutes only up to 4\% of each cycle's period, making energy efficient locomotion an important objective to traverse the required distance of 10 meters per phase. This allows for adequate time to dissipate heat from locomotion power consumption, conduct scientific measurements, generate energy from the solar panels for the next walk sequence, and downlink data to the Earth command/control center. At the same time, the robot requires a stable and safe gait that minimizes dust intrusion into scientific instruments such as the Gravimeter, Ground Penetrating Radar, and Cameras.

\subsection{Reinforcement Learning for legged robot locomotion}
RL formalizes sequential decision making as a Markov Decision Process (MDP), defined by a state space \(\mathcal{S}\), an action space \(\mathcal{A}\), a transition probability distribution \(P(s(t+1)\mid s(t),a(t)\), and a reward function \(R(s,a)\). An agent follows a stochastic policy \(\pi_\theta(a\mid s)\) that maps states to action distributions. The objective is to learn an optimal policy that maximizes the expected cumulative discounted reward \(J(\pi_\theta)\). The expected value under the optimal policy \(\mathbb{E}_{\tau}\) is taken with respect to the trajectory distribution induced by the policy \(\pi_\theta\) and the environment's stochastic dynamics \(P\) \cite{Sutton2020Reinforcementlearning}:

\begin{equation}
    J(\pi_\theta) = \mathbb{E}_{\tau \sim \pi_\theta, P}\!\left[\sum_{t=0}^{\infty} \gamma^t R(s_t,a_t)\right],
    \label{eq:optimal_policy}
\end{equation}
where \(\gamma \in [0,1)\) is the discount factor and \(\tau = (s_0,a_0,s_1,a_1,\dots)\) denotes a trajectory.

Model-free RL does not explicitly model the environment’s state transition dynamics but instead estimates the value function and optimizes the policy gradient by sampling state-action pairs  and fitting them to a latent distribution. In partially observable environments, the policy receives an observation that is mapped from the state \(o(t) = \Omega(s(t)) + \epsilon_t\), where \(\epsilon\) is noise (modeled as Gaussian \(\mathcal{N}(0,\sigma)\)). The policy and value functions are implemented as neural networks, such as a Multi-Layer Perceptron (MLP) with a nonlinear activation function that maps observations to actions or estimates value functions. The MLP is trained to approximate the optimal policy or value function using rewards from sampled state-action pairs. Model-free RL is widely used for locomotion control due to its ability to handle complex environments without modelling dynamics. However, it requires many samples, typically generated through simulation, for the policy to converge.

Differences between the simulated environment used to generate state-action samples and real-world physical conditions can result in a sim-to-real gap, where the trained policies do not generalize well to real-world environments. In the context of complex terradynamic interactions, this poses additional risks, as a policy learned under linear friction conditions is unlikely to generalize to the viscous contacts that occur from foot and granular media interactions. Apart from varying friction, other domain randomization techniques, such as adding random disturbances and observation noise, are omitted, as the study is implemented purely in simulation.

\subsection{Terradynamic Modelling of Robot feet interacting with Granular surfaces}
\label{ssec:terradynamic_model}
Model-based approaches (such as Model Predictive Control) and Model-free approaches for locomotion control (e.g., Reinforcement Learning) both require a physical contact model between the legged robot feet and the ground to accurately predict locomotion behaviour in simulation. Often, these models have idealized assumptions, such as perfectly rigid and inelastic collisions, and use simple foot-ground friction models, such as Coulomb friction or uniform spring-damper physics based on ground penetration \cite{LeLidecEtAl2024ContactModelsRobotics}. The requirement for these models is solver convergence rather than physical accuracy. In Reinforcement Learning, robustness is trained through domain randomization of friction and collision parameters. However, this still includes qualitative assumptions about the physical relationships of these collisions, which do not apply to deformable ground types such as sandy terrain.

These simple, rigid contact models do not accurately capture the terradynamics of robot feet on lunar regolith, a granular substrate similar to sand.  Locomotion control policies trained with these contact models in simulation, despite domain randomization of ground friction parameters, are not robust to instabilities arising from the flow of the soil itself below the feet. These interactions are fundamentally different from the slip due to low friction as modelled in Coulomb friction. Additionally, traversing a granular surface is expected to increase energy and power expenditure, as has been observed in studies on Earth \cite{LiEtAl2010Systematicstudyperformance}.

Contact modelling between rigid and deformable objects is an active field of research, as it impacts many practical robotics tasks such as grasping \cite{SleimanEtAl2023Versatilemulticontactplanning}. The field of locomotion robophysics studies the interactions of robot legs with different media, such as granular substrates \cite{AguilarEtAl2016reviewlocomotionrobophysics}. Motion through granular substrates, such as sand, has been demonstrated to be viscous and nonlinear, depending on the foot shape, angle with the surface, and direction of motion. Early studies aimed at analysing the general properties of rigid/deformable feet and terrain contacts using the Hunt-Crossley model with Coulomb friction \cite{DingEtAl2013Footterraininteractionmechanics}. Since then, numerous studies have attempted to build fundamental models of the terradynamics of legged robots on granular surfaces based on theoretical principles \cite{KangEtAl2018Archimedeslawexplains,AgarwalEtAl2021Surprisingsimplicitymodeling} and experimental data \cite{LiEtAl2013TerradynamicsLeggedLocomotion,AguilarGoldman2016Robophysicalstudyjumping}. 

Terrain identification based gait control was first presented by Wu et al. \cite{WuEtAl2020TactileSensingTerrain}. An alternative approach is to increase robustness by adding domain randomization and disturbances \cite{LeeEtAl2020Learningquadrupedallocomotion}. Recent research by Choi et. al. \cite{ChoiEtAl2023Learningquadrupedallocomotion} integrated a deformable soft ground contact model with Reinforcement Learning control, emphasizing solver convergence and stability, as well as computationally efficient training.

This study implements the terradynamic model from Li et. al. \cite{LiEtAl2013TerradynamicsLeggedLocomotion} with an RL based locomotion policy to investigate the relationship between viscous ground terrain contact dynamics and gait performance. The terradynamic model captures the viscosity of lunar terrain foot-ground interactions, including relationships between the angle of intrusion and motion with the exerted contact forces.  It generalizes to any foot morphology and granular media properties through simple coefficients that can be identified through experiments or inferred from empirical relationships.

This model was built experimentally by intruding a thin and rigid plate moving through the granular medium. The vertical and horizontal stresses \(\sigma_z\) and \(\sigma_x\) are measured for small segments on the plate, capturing their relationship with the angle of attack to the medium layer \(\beta_s\), the angle of intrusion (direction of motion) \(\gamma_s\), and the penetration depth \(|z|_s\). The lift and drag forces \(F_z\) and \(F_x\) on the foot shape are then determined as the integration of stresses acting on the leading surface area S intruding into the granular medium with segment area \(dA_s\):

\begin{equation}
    F_{z,x} = \int_S{\sigma_{z,x}(|z|_S,\beta_S,\gamma_S) dA_s}.
    \label{eq:terrain_contact_model}
\end{equation}

The experiments found that the stress models for all granular media tested generalized to a model with 9 zero and first order Fourier coefficients for \(\sigma_{z,x}(\beta_S,\gamma_S)\) linearly multiplied with the penetration \(z_s\), and scaled with a factor \(\zeta\) that captures the compactness and grain size effects for the medium:

\begin{align}
    \sigma_z &= \zeta |z_s|\sum_{m=-1}^{1} \sum_{n=0}^{1}\left[ A_{m,n} \cos(2m\beta+n\gamma) + B_{m,n} \sin(2m\beta+n\gamma)\right],\\
    \sigma_x &= \zeta |z_s|\sum_{m=-1}^{1} \sum_{n=0}^{1}\left[ C_{m,n} \cos(2m\beta+n\gamma) + D_{m,n} \sin(2m\beta+n\gamma)\right].
\end{align}

The generic Fourier coefficients used are summarized in Table \ref{tab:terrain_model_coefficients}. 

\begin{table}[!ht]
\centering
\caption{Generic Fourier coefficients for the terradynamic stress model by Li et. al. \cite{LiEtAl2013TerradynamicsLeggedLocomotion}, where values are in \(\text{N}.\text{cm}^{-3}\) for penetration depth \(|z|_S\) measured in cm.}
\label{tab:terrain_model_coefficients}
    \begin{tabular}{ccclcl}\toprule
         \({m,n}\)  &  0,0&  1,0&  0,1&1,1 &$-$1,1\\\midrule
         \(A_{m,n}\)&  0.206&  0.169& -  & - & - \\
         \(B_{m,n}\)& - & - &  0.358&0.212&0.055\\
         \(C_{m,n}\)& - & - &  0.253&-0.124&0.007\\
         \(D_{m,n}\)& - & - 0.088&  - & - & - \\ \bottomrule
    \end{tabular}
\end{table}
This formulation makes the model suitable for domain randomization by varying the stress coefficient \(\zeta\) in the RL algorithm, which helps generalize the gait to variations in lunar regolith material properties. Furthermore, the experimental approach can be used to find Fourier coefficients for lunar regolith equivalent materials and build a more accurate model for the granular surface material.
\section{Method \& Implementation}
\label{sec:methods}

\subsection{Terrain Modelling}
\label{ssec:terrain_modelling}
The implementation of the terradynamic model described in section \ref{ssec:terradynamic_model} is generalized for any foot morphology through a discrete mesh representation. The faces of the mesh are taken as the finite area segments in the calculation of the lift and drag forces \eqref{eq:terrain_contact_model}. These segment forces are then summed over the body of the foot to calculate the total force and torque acting on the origin of the foot frame, and applied as an external force and external torque in the simulation environment. The implementation generalizes the formulation by Li et. al. \cite{LiEtAl2013TerradynamicsLeggedLocomotion} to use the ground normal, thereby enabling its use on non-flat terrains by projecting the segment frames with respect to the ground normal and scaling the surface depth penetration with a gravity-angle-based scaling factor.

\begin{figure}[!hb]
    \centering
    \includegraphics[width=\linewidth,trim=0.5cm 2.5cm 0.5cm 1.5cm, clip]{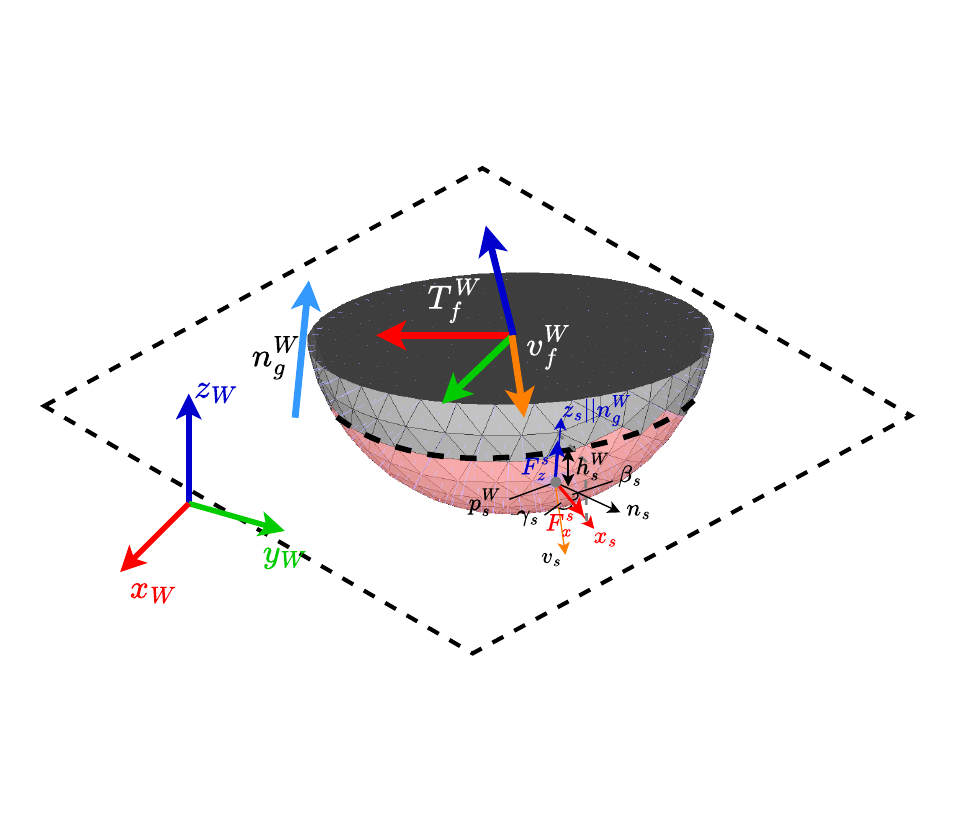}
    \caption{Terradynamic contact model, showing a hemispheric mesh foot intruding into the surface layer with height \(z_h\). The contacting foot segments are shown in red. The world frame orientations are shown with coloured axes, and velocities in orange.}
    \label{fig:terrain_model}
\end{figure}

The following initialization step is carried out: the mesh faces are parsed into segments indexed \(s\) with foot body frame centroids \(\vect{c}^B_s \in \nR{3}\), normals \(\vect{n}^B_s \in \nR{3}\) and areas \(dA_s \in \nR{}\). Then, for every simulation step, the cumulative forces and torques acting on foot \(f\) are calculated. At a high level, 
the algorithm projects segment coordinates between body and world frames, checks for surface penetration and leading surfaces, calculates \(\beta_s\) and \(\gamma_s\), computes the stresses using \eqref{eq:terrain_contact_model}, transforms forces back to the world frame, and sums them to calculate the cumulative force and torque acting on each foot. The detailed implementation is elaborated in Algorithm \ref{alg:terrain_contact}. A figure illustrating the relevant parameters in the Algorithm is shown in Figure \ref{fig:terrain_model}. 

\begin{algorithm}[!ht]
\caption{Terradynamic Contact Model}
\label{alg:terrain_contact}
% \small
\begin{algorithmic}[1]
\For{each foot \( f \in F \) }
    \State Get the foot world frame pose \(\vect{T}^W_f = \begin{bmatrix}\vect{R}^W_f & \vect{p}^W_f \\ \vect{0}_{1 \times 3} & 1\end{bmatrix}\), linear velocity \(\vect{v}_f^W\) and angular velocity \(\vect{\omega}^W_f\)
    \State Initialize total foot force \(\vect{F}^W_f \gets \vect{0}\) and torque \(\vect{\tau}^W_f \gets 0\)
    \For{each segment \( s \in S\) with body frame normal \( \vect{n}^B_s \), centroid \( \vect{c}^B_s \), area \( A_s \)}
        \State Calculate the normal and centroid in the world frame: \quad \( \vect{n}^W_s \gets \vect{R}_f^W \vect{n}^B_s\), \quad \( \vect{c}^W_s \gets \vect{T}_f^W \vect{c}^B_s\)
        \State Get ground height at segment position \(z_g^W\), and the ground normal at segment position \(\vect{n}_{g}^W\)
        \State Calculate surface penetration of into the ground: \(z^W_s = -\icol{0 & 0 & c_s^W - z_g^W} \cdot \vect{n}_{g}^W\)
        \State Surface penetration and leading segment check: \textbf{if} \(z^W_s > 0\) skip segment
        \State Calculate segment velocity in the world frame: \( \vect{v}^W_s \gets \vect{v}^W_f + \vect{c}_{sf}^W \times \vect{\omega}_f^W\)
        \State Leading segment check: \textbf{if} \(\vect{v}_s^W \cdot \vect{n}_s^W < 1\poft{-6}\) skip segment
        \State Align segment frame with ground plane: \(\vect{n}_s^g = \vect{n}_s^W - \left(\vect{n}_s^W \cdot \vect{n}_g^W \right) \vect{n}_g^W\)
       % \State Set direction: \parbox[t]{\dimexpr\textwidth-2\algorithmicindent}{\scriptsize
       %      \( \hat{\mathbf{d}}^W_s \gets \begin{cases} 
       %  \frac{\vect{n}_s^g}{\|\vect{n}_s^g\|} & \text{if } \|\vect{n}_s^g\| < 10^{-9} \\
       %  \begin{cases} 
       %      (1, 0, 0) & \text{if } \|\mathbf{v}_{xy}\| < 10^{-9} \\
       %      -\frac{\mathbf{v}_{xy}}{\|\mathbf{v}_{xy}\|} & \text{otherwise}
       %  \end{cases} & \text{otherwise}, \ \mathbf{v}_{xy} = (v_s(x), v_s(y), 0)
       %  \end{cases} \)}
       \State Calculate rotation from world frame to ground-aligned segment frame: \(\vect{R}^W_{ws} \gets \icol{\vect{n}_s^W & \vect{n}_s^g \times \vect{n}_s^W  & \vect{n}_g^W}\)
       \State Project foot normal and velocity into ground-aligned segment frame: \(\vect{n}_s^g \gets \vect{R}^W_{ws} \vect{n}^W_s\), \(\vect{v}_s^g \gets \vect{R}^W_{ws} \vect{v}^W_s\)
        \State Compute angle of intrusion and angle of incidence: 
        \quad \( \beta_s \gets \arctan(n_s^s(x), -n_s^s(z)) \), 
        \quad \( \gamma_s \gets \begin{cases} 
            0 & \text{if } |-v_s^s(z)| < 10^{-9} \\
            \arctan(v_s^s(x), -v_s^s(z)) & \text{otherwise}
        \end{cases} \)
        \State Calculate the effective surface depth adjusting for gravity acting along the ground normal: \(z_s^g \gets z_s^g | \irow{0 & 0 & 1} \vect{n}^W_g |\)
        \State Compute segment forces: \quad \( F_z^s \gets \sigma_z(z_s^g, \beta_s, \gamma_s) dA_s\) \quad \( F_x^s \gets \sigma_x(z_s^g, \beta_s, \gamma_s) dA_s \)
        \State Total force in the segment frame: \( \vect{F}_s^s \gets \icol{F_x^s, 0, F_z^s} \)
        \State Reproject segment force to the world frame: \( \vect{F}^W_s \gets \vect{R}_{ws}^\top \vect{F}_s^s \)
        \State Accumulate total foot force: \( \vect{F}^W_f \gets \vect{F}_f + \vect{F}_s^W \)
        \State Accumulate total foot torque: \( \vect{\tau}_f \gets \vect{\tau}_f + ( \vect{c}^W_s - \vect{p}^W_f) \times \vect{F}_s^W \)
    \EndFor
    \State Apply external force \( \vect{F}_f \) and external torque \( \vect{\tau}_f \) to foot F at \(\vect{p}^W_f\)
\EndFor
\end{algorithmic}
\end{algorithm}

\subsection{Locomotion Gait}

The locomotion control was implemented through training a Deep Reinforcement Learning policy trained using Proximal Policy Optimization (PPO) \cite{2017ProximalPolicyOptimization}, a variant of the Actor-Critic method that updates the network weights by maximizing a clipped parameter surrogate objective to ensure stable gradient descent using the ADAM optimizer. 
%It uses an actor-critic architecture \cite{Sutton2020Reinforcementlearning}, where the actor is the policy network with imperfect and physically measurable set of observations \(o^\pi\), while the critic is a network with privileged observation states \(o^V_t\). The critic provides a more accurate value function estimate \(V_\phi(o^V_t)\) in the form of an estimated advantage function to the actor \(\hat{A}_t=\hat{Q}^\pi(o^V_t,a_t)-\hat{V}^\pi(o^V_t)\). The actor uses this information to adjust the policy to increase the expected reward \eqref{eq:optimal_policy}, which accelerates convergence to the optimal policy. 

The observations, rewards, actions, neural network, and training are similar  to \cite{ArmEtAl2025EfficientLearningBased}, with the following small modifications:
\begin{enumerate}
    \item Note that the power penalty, which uses the power loss equations \eqref{eq:power_loss}, is composed of \(\vect{\tau}^2\) and power  \(\vect{\tau}\dot{\vect{q}}\) components with coefficients. Therefore, these are set as torque and mechanical power penalties, with the weights re-tuned to result in a stable locomotion gait. This allows us to decouple the motor performance characteristics from the RL policy training.
    \item A small penalty for deviating from a flat base orientation is added, which improves convergence to a gait where the base is stable, particularly in the granular contact environment where slip forces on the foot can easily imbalance the robot.
    \item Disturbances in velocity, applied forces, and observation noise are omitted to speed up policy training, as they are not required for this analysis.
    \item Curriculum training is applied through a linear scaling up of terrain difficulty from the episode when it is started.
\end{enumerate}

 Tracking is done via a base-frame velocity command, composed of the forward and sideways linear velocities \(v^B_{c,x}\), \(v^B_{c,y}\), as well as a yaw turn rate \(v^B_{c,\psi}\) about the base centre, provided as an observation. The full set of observations is shown in Table \ref{tab:lunarleaper_observations}. 
 
 The output of the policy is actions \(\vect{a} \in \nR{12}\) that map to target joint angles via a PD controller \(\vect{q}_{*} = 0.3 \vect{a} + \vect{q}_i\), where \(\vect{q}_i\) are the default configuration joint angles for the quadruped robot model.
Both the actor and critic networks are Multi-Layer Perceptrons with 3 fully connected layers of dimensions [512, 256, 128], using Exponential Linear Unit (ELU) activation functions at each layer. 

\begin{table}[!hb]
\centering
\caption{Observation inputs used in the actor and critic networks for training the RL policy.}
\label{tab:lunarleaper_observations}
    \begin{tabular}{llll}\toprule
    \textbf{Observation}&  \textbf{Dimension} &\textbf{Actor/Critic}& \textbf{Formulation}\\\midrule
    Base Linear Velocity& 3&Critic& \(\vect{v}_B(t)\)\\
    Previous Angular Velocities& 21& Actor&\(\icol{\omega_B(t-7dt) & ... & \omega_B(t-dt)}\)\\
    Base Angular Velocity& 3& Both&\(\omega_B(t)\)\\
    Projected Gravity  & 3   &Both& \({R^W_B}^\top \icol{0 & 0 & -1}\) \\
    Velocity Command   & 3   &Both& \(0.5\icol{v^B_{c,x} & v^B_{c,y} & v^B_{c,\psi}}\)\\
    Joint Angles       & 12  &Both& \(\vect{q}\) \\
    Joint Velocities   & 12  &Both& \(\dot{\vect{q}}\)\\ 
    Last Action        & 12  &Both& \(\vect{a}_{t-1}\) \\ \bottomrule
    \end{tabular}
\end{table}

The Rewards were split into two categories: tracking rewards, which encourage convergence to a stable walking gait according to the commanded linear velocity, and smoothing rewards, which reduce jerky motions, prevent high joint torques, and reduce power consumption \cite{ArmEtAl2025EfficientLearningBased}. Initially, the training begins with only tracking rewards to encourage exploration in the region of feasible walking gaits. After 200 episodes of simulation rollouts, the smoothing rewards are introduced gradually, with a linear scaling factor applied across 50 episodes. The full set of rewards is shown in Table \ref{tab:lunarleaper_rewards}.

\begin{table}[!ht]
\centering
\caption{Rewards used to train the locomotion and base pose controllers. The tracking error is in the base frame calculated from the commanded and actual linear velocities \(\icol{v^B_x & v^B_y & v^B_\psi}\), as illustrated for the forward: \({e}_{v,x} = v^B_{c,x}-v^B_x\).}
\centering
\label{tab:lunarleaper_rewards}
    \begin{tabular}{lll}\toprule
         \textbf{Reward}&  \textbf{Equation}& \textbf{Weight}\\ \midrule
 \textbf{Tracking Rewards}& &\\\midrule
         Linear velocity tracking&  \(\exp\{-(e^2_{v,x}(t)+e^2_{v,y}(t))\}/0.25\)& 1.0\\
         Yaw velocity tracking&  \(\exp\{-e_{\psi}^2\}/0.25\)& 0.5\\
         Flat orientation& \(\icol{0 & 0 & -1}-{R^W_B}^\top \icol{0 & 0 & -1}\) & -0.2\\
         Joint Limit Violation& \(\|\max(0, \vect{q}^{\min} - \vect{q}(t)) + \max(0, \vect{q}(t) - \vect{q}^{\max}) \|_2^2\)& -1.0 \\
 Undesired contact& \(n_c\)&-1.0\\
         Foot impact velocity& \({v^W_{f,z}}^2\) on contact& -0.6 \\\midrule
 \textbf{Smoothing Rewards}& &\\ \midrule
 Joint Torque& \(\|\vect{\tau}(t)\|^2\)& \(-1\poft{-5}\) \\
 Joint Acceleration& \(\|\ddot{\vect{q}}(t)\|^2\) & \(-1\poft{-7}\)\\
 Action Rate& \(\|\vect{a}(t)-\vect{a}(t-dt)\|^2\)& \(-0.08\)\\
 Total Power& \(\|\vect{\tau}(t)^\intercal\dot{\vect{q}}(t)\|^2\) & \(-1\poft{-6}\)\\ \bottomrule
    \end{tabular}
\end{table}
The RL policy training parameters and features are kept the same for both rigid and granular contact environments to allow for a fair comparison of the locomotion performance.

\section{Results}
\label{sec:results}
\subsection{Simulation environment}
The RL training environment with the terradynamic model was implemented in RaiSim \cite{raisim}, which allows for low-level calculation and application of external forces with parallelization capabilities. Two types of environment were implemented: one using the standard Rigid contact Raisim solver using Coulomb friction for foot-terrain interactions, and a second `soft' environment implementing the terradynamic model outlined in section \ref{ssec:terrain_modelling}. The soft environment relied on the terradynamic contact model, implemented as penetration with a virtual soil layer of height \(z_H=0.05\)m, with the rigid ground as a fallback for deep extrusions.

The legged robot model used was a simplified analogue of Magnecko \cite{LeuthardEtAl2024MagneckoDesignandControl}, with a uniform density base of 4.5 kg, lightweight aluminium links with a mass density of \(0.2\)kg/m, and motors corresponding to the mass/inertial characteristics of the Maxon DT50M, resulting in a total mass of 13.5 kg. The rigid terrain training used a spherical foot model where only the bottom hemisphere made contact with the ground, as the solver is more stable and faster in computation for rolling point contacts. The soft terrain feet were modelled as a hemisphere discretized with 688 faces. 

Finding the right scaling factor for the calculation of segment stresses is of particular importance, as it affects the slip, sink-in, and elastic collision properties of the terradynamic interaction. The supplementary material of \cite{LiEtAl2013TerradynamicsLeggedLocomotion} provides the scaling factor of closely packed poppy seeds as 0.488. The stress values are given in N\(\cdot\text{cm}^3\) and are multiplied by depth \(|z|_S\) which is in cm. To convert this into SI units, the scaling factor is turned into a stress density, which multiplies the scaling factor by \(1\poft{8}\). This was scaled by another two first-order factors: earth to moon gravity (\(1.62/9.81 = 0.165\)), and the grain size and compactness ratio of poppy seeds (580 kg\(\cdot \text{m}^3\) \cite{SacilikColak2005DielectricPropertiesOpium}) to JSC1A (1500 kg\(\cdot \text{m}^3\)), a Lunar regolith simulant \cite{Kovtun2023OverviewLunarRegolith}. The resulting stress density used is \(2.15\poft{7}\). Studies of depth penetration in granular media show that this is an acceptable first-order approximation \cite{BrzinskiEtAl2013DepthDependentResistance,DacaEtAl2023ExpansionExperimentalEvaluation}. For comparison with the rigid contact model, the static and dynamic friction was set to 0.3, simulating a slippery surface as similar as possible to the soft contact slip.

Two types of domain randomization are implemented: friction and terrain through curriculum training, which are randomized separately for every parallel environment. For the rigid contact environment, the Static Coulomb friction is uniformly sampled between \(\irow{0.3, 1.0}\) and the dynamic Coulomb friction set as a factor of the static friction, uniformly sampled in the interval \(\irow{0.3, 1.1}\) with a minimum of 0.1. For the soft contact environment, the stress density was set uniformly to \(2.15 \poft{7}\). 

The terrain curriculum training used procedural generation of a rough and undulating surface using Perlin randomization, starting with a flat surface from episode 1000 for the rigid contact environment and episode 1100 for the soft contact environment, updated every 50 episode with a difficulty scale factor of \(0.75\poft{-3}\) per episode. This mostly simulated a gentle slope, as the soft contact model would break down and result in abnormally high contact forces at high angles, causing the robot to fly up in the air, which resulted in unstable training and divergence from a feasible gait.

The RL training was conducted with 1000 parallel simulation environments, with each episode lasting 20 seconds, sampled at a controller frequency of 50 Hz. The simulation solver was set to 400 Hz for the rigid contact environment and 800 Hz for the soft contact environment, as the soft contact forces were more prone to numerical instabilities and variations. The velocity commands given to the observation inputs in every environment were uniformly randomly sampled between \([-1.0, 1.0]\) for \(v^B_{c,x}\) and \(v^B_{c,x}\), and \([-0.75, 0.75]\) for \(\omega_{c,\psi}\). New commands were given every 10 seconds of simulation time, i.e., twice in an episode. 

The RL policy training was conducted on 44 GB of CPU memory split across 36 cores and a single 4 GB memory GPU for updating the RL networks. As the soft contact environment required a higher solver frequency and more operations over the discretized mesh compared to the rigid environment, the training time was, on average, doubled per episode. The learning parameters are summarized in Table \ref{tab:rl_training_parameters}.

\begin{table}[!h]
    \centering
    \caption{Parameters for the RL Training.}
    \label{tab:rl_training_parameters}
    \begin{tabular}{ll}\toprule
         \textbf{Parameter}& \textbf{Value} \\\midrule
         Discount Factor \(\gamma\) & 0.99\\
         Entropy Coefficient & 0.007\\
         Learning Rate & \(1\poft{-3}\)\\
         Max Gradient Norm & 0.95\\
         Generalized Advantage Estimate \(\lambda\) & 0.95\\
         Number of mini-batches & 4\\
         Number of learning episodes & 4\\
         Clip Range & 0.2\\
         Total episodes & 2000\\ \bottomrule
    \end{tabular}

\end{table}

The power reward described in Table \ref{tab:lunarleaper_rewards} used the mechanical power \(P_{mech}(t) = \tau \dot{q}\) for the torque \(\tau\) and \(\dot{q}\) velocity for each motor, assuming no power recovery (i.e., the minimum is 0). However, in reality, there are significant losses from the winding resistance and gears depending on the torque/velocity regions in which the motor operates. To account for these, during the analysis of the locomotion policy, the power losses are calculated according to the power loss model in \eqref{eq:power_loss}:

\begin{align}
    P_{loss}(t)    &= P_{gear}(t) + P_{motor}(t) + P_{winding}(t) + P_{driver}(t)\label{eq:power_loss},\\
    P_{motor}(t)   &= (1-\eta_{motor}) P_{mech}(t),\\
    P_{gear}(t)    &= (1-\eta_{gear}) P_{mech}(t),\\
    P_{driver}(t)  &= \frac{\alpha}{(\eta_{gear}N \kappa_\tau)^2}\tau(t)^2 + \frac{\beta}{\eta_{gear}\kappa_\tau} P_{mech} + P_{elec},\\
    P_{winding}(t) &= \frac{3 R_s}{(\eta_{gear}N \kappa_\tau)^2} \tau(t)^2,
\end{align}

where \(\kappa_\tau\) is the motor torque constant, \(N\) is the gear ratio, \(\eta_{gear}\) and \(\eta_{motor}\) are gear and motor efficiency factors, respectively, \(R_s\) is the winding resistance per phase, \(\alpha_1\) and \(\alpha_2\) are generic coefficients, and \(P_{elec}\) is the constant electric loss. The total power is \(P_{total}(t) = P_{mech}(t) + P_{loss}(t)\). The values used for the power loss coefficients are summarized in Table \ref{tab:motor_model_coeff}. The parameters for which the values from the motor datasheet can be extracted are included in the model, and the unknown parameters are set to 0.

\begin{table}[!ht]
    \centering
\caption{Motor model parameters for the Maxon DT50M used used in the power loss calculations.}
\label{tab:motor_model_coeff}
    \begin{tabular}{lll}\toprule
         \textbf{Parameter} &  \textbf{Unit} & \textbf{Value}\\\midrule
         \(\eta_{gear}\) &  -  & 0.95\\
         \(\eta_{motor}\)&  - & 0.92\\
         \(R_s\)         &  \(\Omega\) & 0.748\\
         N               & -  &18\\
         \(\kappa_\tau\)&  Nm/A & 0.105\\ 
         \(\alpha\) & - & 0.0 \\
         \(\beta\)  & - & 0.0 \\ 
         \(P_{elec}\) & W & 0.748 \\\bottomrule
    \end{tabular}
\end{table}

\subsection{Terrain model and locomotion gait comparison}
After training the policy, two types of analyses were conducted: a qualitative analysis of the robot locomotion gait and a quantitative analysis of the torque/power profiles. The qualitative analysis involved a visual inspection of the gaits to classify their types (walk, trot, pronk, etc.) and characteristics in terms of stride length, height, and foot contact/air time. This followed an iterative process to tune the rewards and produce a consistent, periodic, and energy efficient gait.

A challenge in reward tuning for the gait was finding a common parameter set that resulted in stable training convergence to a walking gait for both the rigid and soft contact models. Typically, the soft contact model was more restrictive, so after tuning the rigid contact walk, the same reward parameters were tested in the soft terrain model, for which there was a failure to generalize to the highly nonuniform and nonlinear dynamics of the foot-terrain contacts. The number of MLP network layers and their size also affected the ability of the network to learn a gait for the soft terrain.

Training a soft contact policy from scratch purely in the soft contact environment did not result in a smooth or feasible gait, as early in the training the contacts were highly unstable, which resulted in convergence to overly conservative gaits with small scuttle steps. For this reason, the soft contact model used transfer learning, taking a rigid model trained for 800 episodes (where it learns to stand and perform basic walking with wide stride steps) as the base model to start training from 0 episodes.

Nevertheless, the qualitative nature of the gaits differs between rigid and soft contact environments. The rigid contact gait uses wider strides as the gait is more confident of a high impulse contact reaction that helps it establish a foothold. The soft contact gait can have significant sideslip with such high foot velocity impacts, which increases the tracking error. Hence, the strides taken with a soft contact model are smaller and have a higher duty cycle to minimize slip forces, which are proportional to the foot-surface contact angle and direction of motion. 

Unlike the rigid contact model, which assumes no ground force being exerted during the upswing of a gait, the soft terrain model has intrusion as well as extrusion forces, albeit at a lower magnitude. This, coupled with the variance in drag forces from different angles of attack and intrusion, results in degraded tracking, as it is harder to achieve a stable foothold in granular media. With 2000 episodes of training, the gait evened out in terms of a net tracking error of 5\% over a 10 m traversal.

\begin{figure}[!ht]
    \centering
    \includegraphics[width=0.8\linewidth]{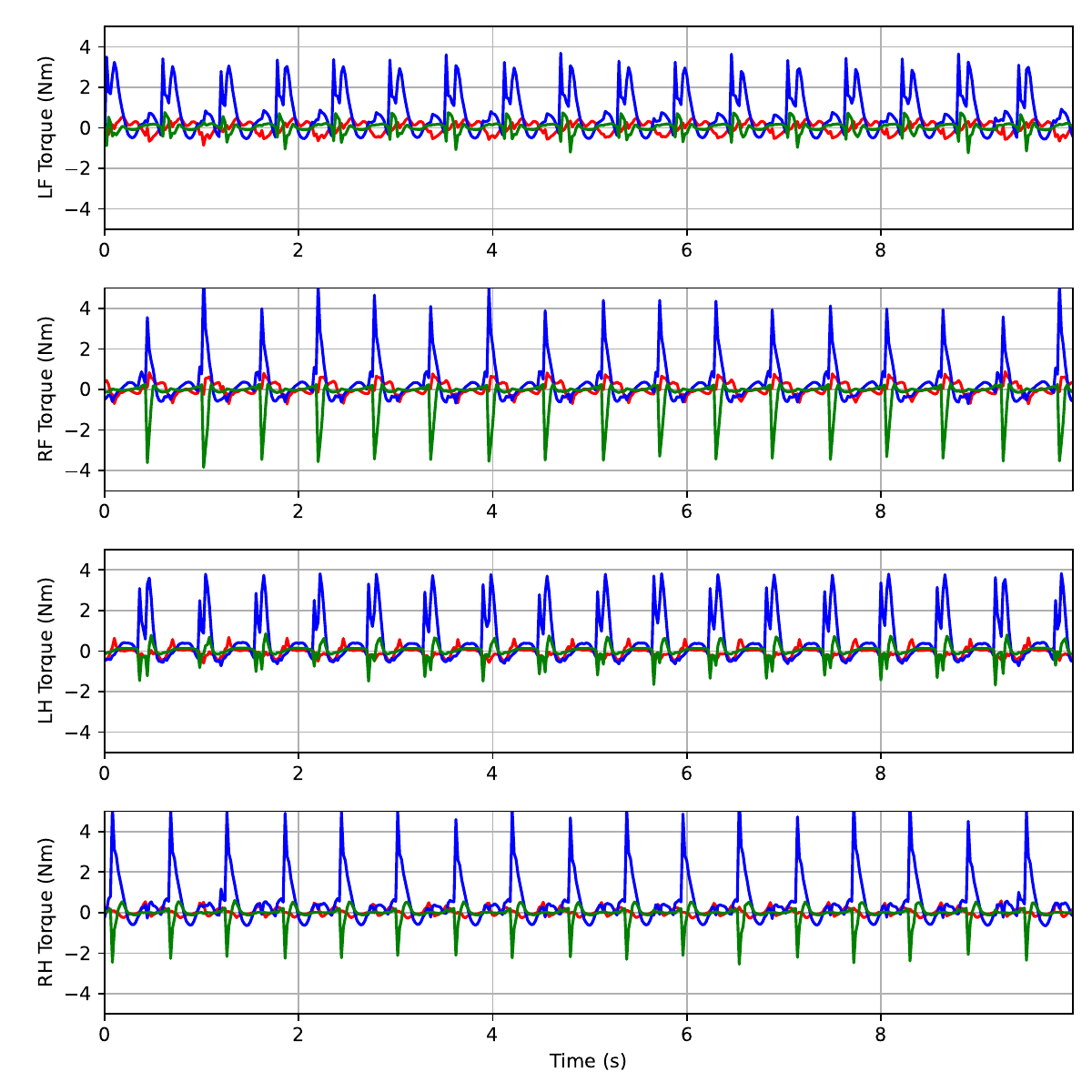}
    \caption{Torques for slow walk (0.2m/s), rigid terrain, showing yaw (red), pitch (blue) and knee (green) joints.}
    \label{fig:rigid_walk_bl_slow_torques}
\end{figure}

\begin{figure}[!h]
    \centering
    \includegraphics[width=0.8\linewidth]{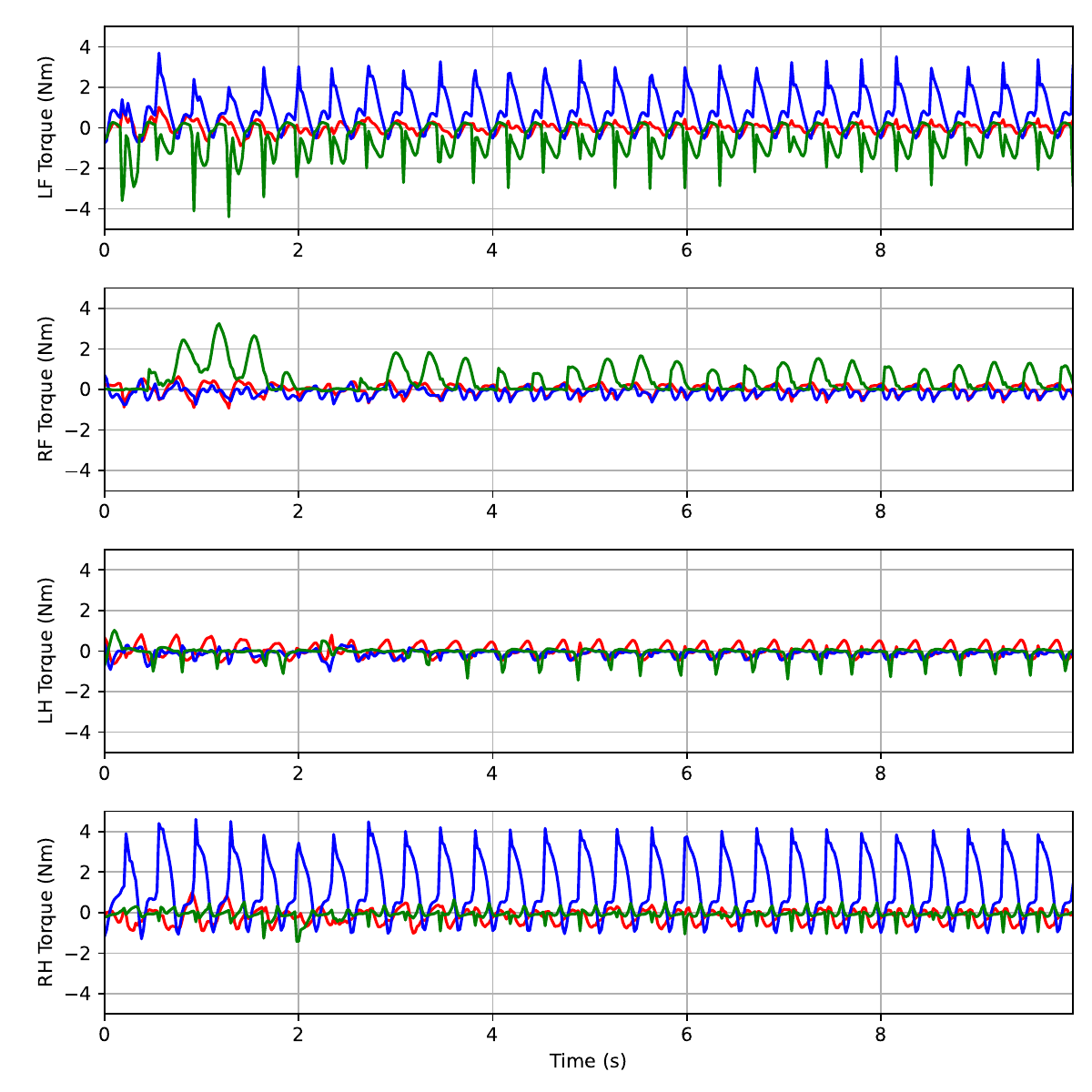}
    \caption{Torques for slow walk (0.2m/s), granular contacts, showing yaw (red), pitch (blue) and knee (green) joints.}
    \label{fig:soft_walk_bl_slow_torques}
\end{figure}

\begin{figure}[!h]
    \centering
    \begin{subfigure}[b]{\textwidth}
        \centering
        \includegraphics[width=0.8\linewidth]{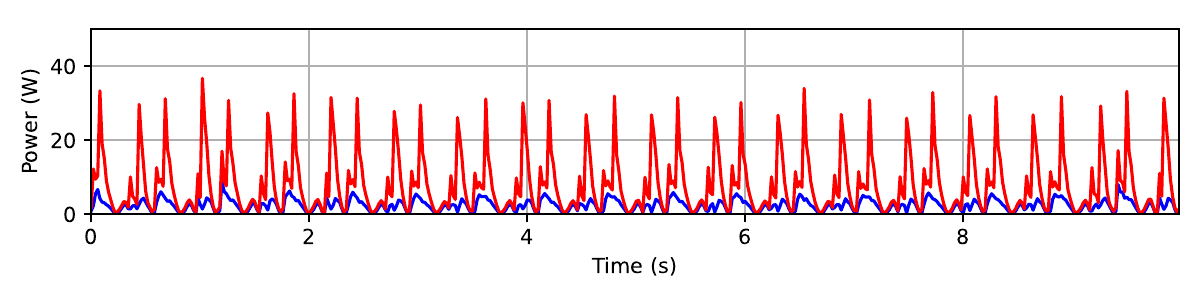}
        \caption{Rigid contacts}
    \end{subfigure}
    \hfill
    \begin{subfigure}[b]{\textwidth}
        \centering
        \includegraphics[width=0.8\linewidth]{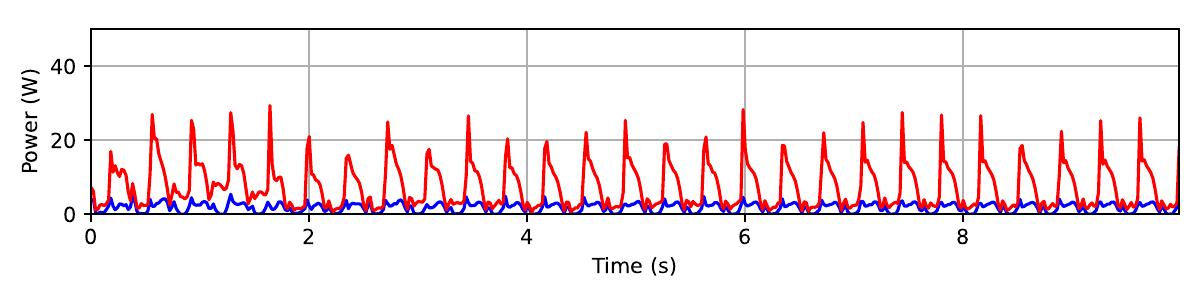}
        \caption{Soft contacts}
    \end{subfigure}
    \caption{Mechanical (blue) and total (red) Power consumption for slow walk (0.2m/s).}
    \label{fig:slow_walk_power}
\end{figure}

\begin{figure}[!h]
    \centering
    \includegraphics[width=0.8\linewidth]{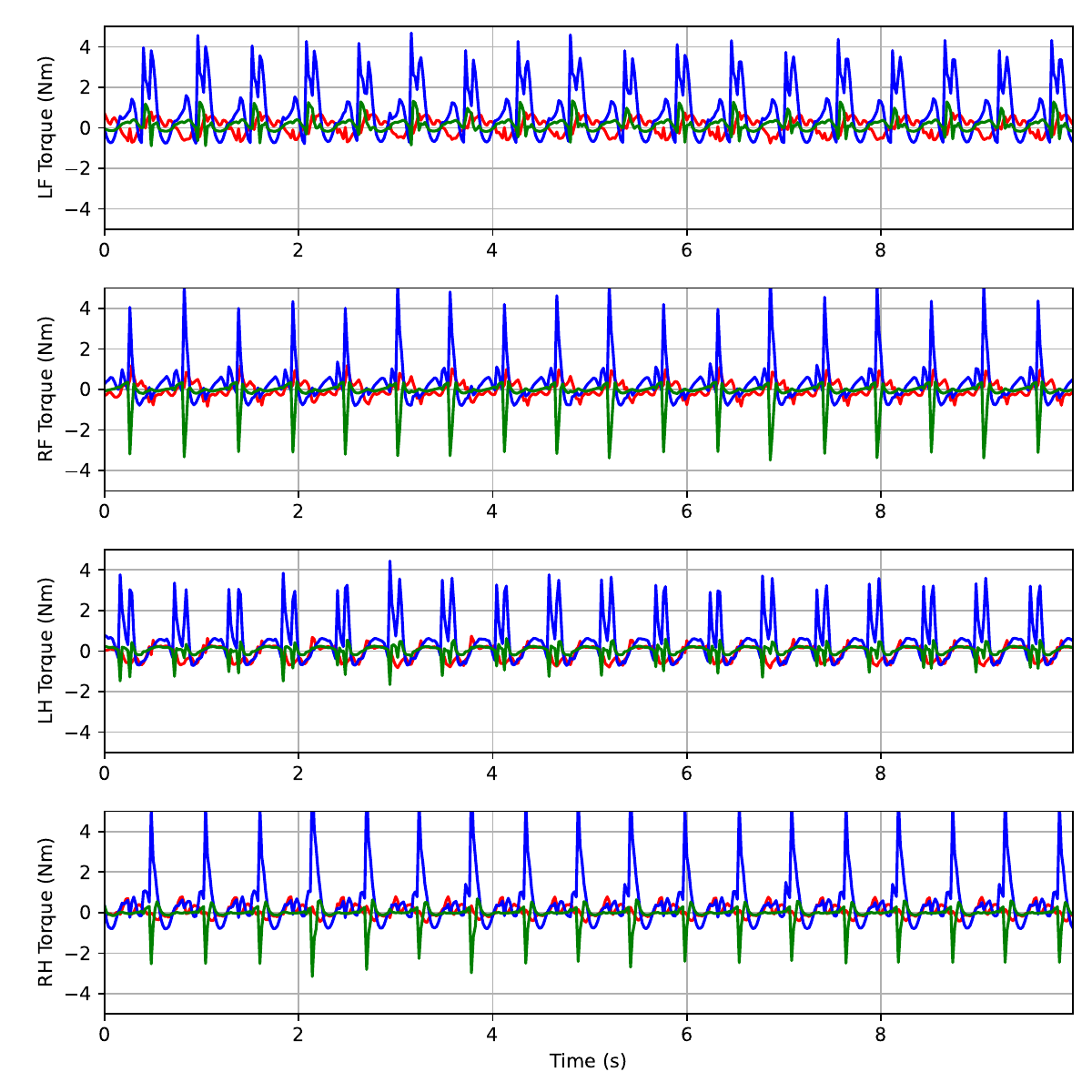}
    \caption{Torques and Power for fast walk (0.4m/s), granular contacts, showing yaw (red), pitch (blue) and knee (green) joints.}
    \label{fig:rigid_walk_bl_fast_torques}
\end{figure}

\begin{figure}[!h]
    \centering
    \includegraphics[width=0.8\linewidth]{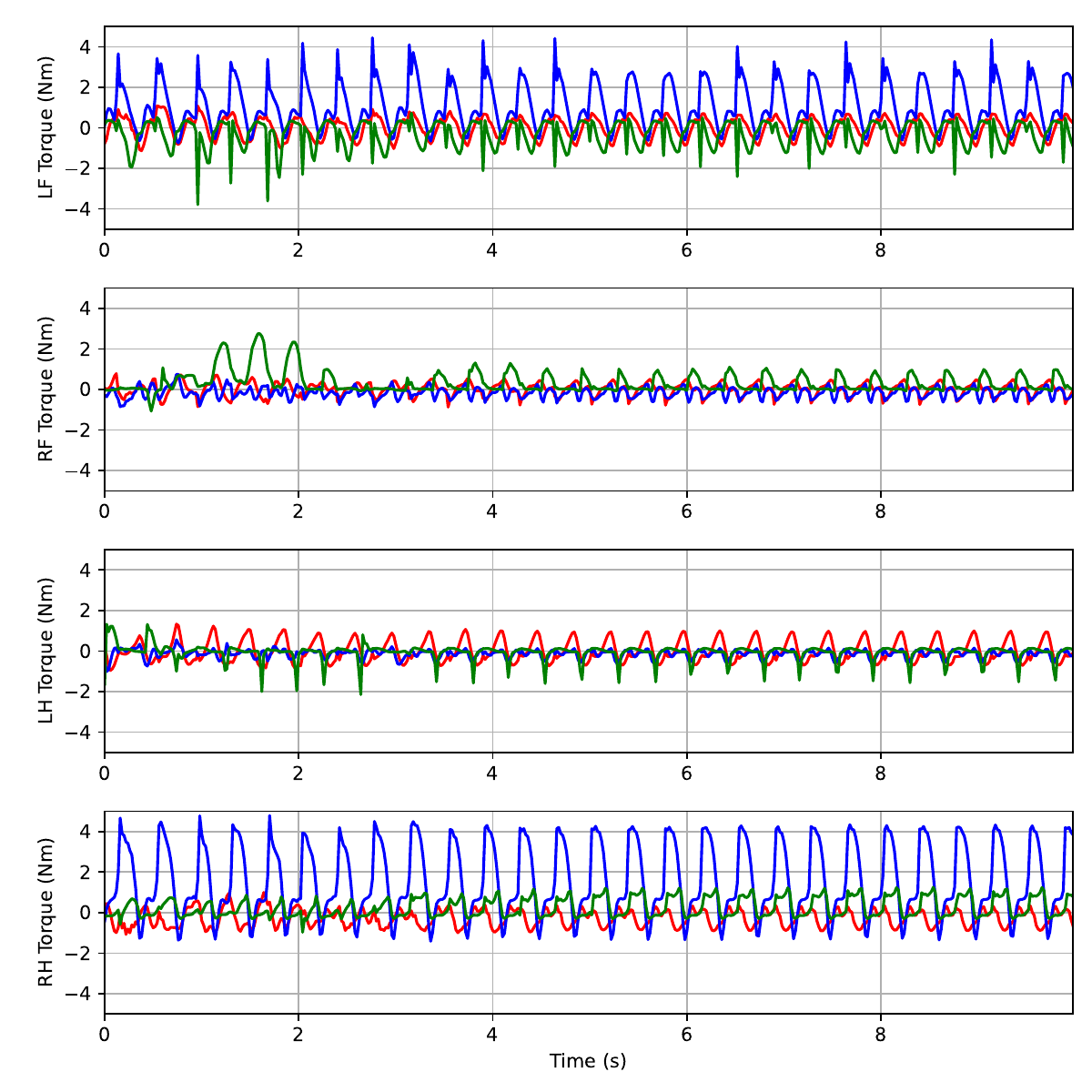}
    \caption{Torques and Power for fast walk (0.4m/s), granular contacts , showing yaw (red), pitch (blue) and knee (green) joints.}
    \label{fig:soft_walk_bl_fast_torques}
\end{figure}

\begin{figure}[!h]
    \centering
    \begin{subfigure}[b]{\textwidth}
        \centering
        \includegraphics[width=0.8\linewidth]{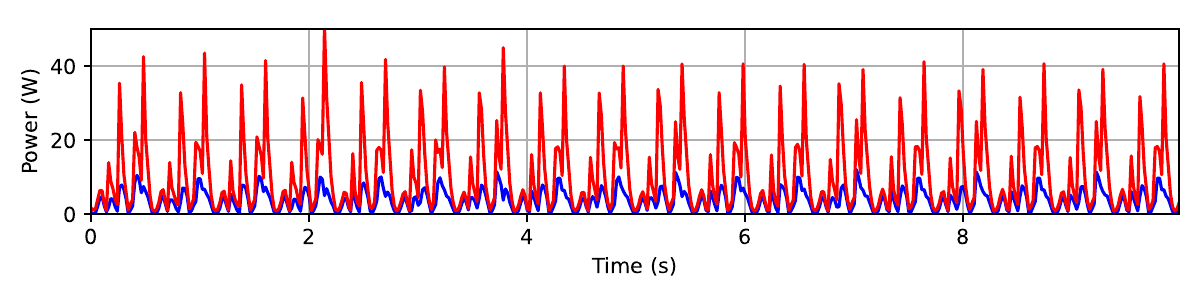}
        \caption{Rigid contacts}
    \end{subfigure}
    \hfill
    \begin{subfigure}[b]{\textwidth}
        \centering
        \includegraphics[width=0.8\linewidth]{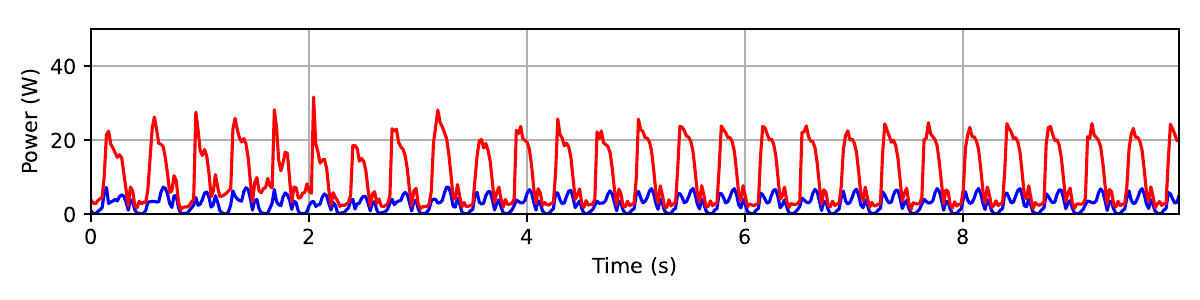}
        \caption{Soft contacts}
    \end{subfigure}
    \caption{Mechanical (blue) and total (red) Power consumption for fast walk (0.4m/s).}
    \label{fig:fast_walk_power}
\end{figure}

The torque and power profiles for a flat walk are presented in Figures \ref{fig:rigid_walk_bl_slow_torques} and \ref{fig:soft_walk_bl_slow_torques} for 0.2 m/s and in Figures \ref{fig:rigid_walk_bl_fast_torques} and \ref{fig:soft_walk_bl_fast_torques} for 0.4 m/s under the rigid and soft terrain contact models. The power consumption comparison for the same walks is shown in Figures \ref{fig:slow_walk_power} and \ref{fig:fast_walk_power}.  

To measure aggregate statistics over variations in different walk directions, parallel simulations of different traversals were carried out. Two types of traversals were benchmarked: a 50 second walk at 0.2 m/s (slow) and a 25 second walk at 0.4 m/s (fast), both covering a total of 10 m in each walk, each on a gently sloped terrain. Uniform random variations in the side velocity of up to 50\% with yaw velocity components of up to \(\pm 0.05\)rad/s were added to introduce diversity in the traversals. These results are summarized in Tables \ref{tab:lunarleaper_traversal_slow} and (insert). The absolute joint torques, mechanical power, power loss, total energy, as well as the Mechanical Cost of Transport (MCOT) are shown.

\begin{table}[!hb]
\caption{Aggregate locomotion statistics across 100 traversals of 50 second segments at a slow walk of 0.2m/s, showing mean (\(\mu\)), standard deviation (\(\sigma\)), and maximum values.}
\label{tab:lunarleaper_traversal_slow}
\centering
\begin{tabular}{@{}lllll@{}}
\toprule
\textbf{Metric}            & \multicolumn{2}{l}{\textbf{Rigid terrain}} & \multicolumn{2}{l}{\textbf{Soft terrain}} \\ \midrule
                     & \(\mu\pm\sigma\)   & Max  & \(\mu\pm\sigma\)    & Max   \\ \midrule
Absolute Joint Torque (Nm) & \(0.418 \pm 0.685\)& 7.33& \(0.521 \pm  0.688\)& 5.97\\
Mechanical Power (W) & \(2.27 \pm 2.04\)& 13.8& \(2.01 \pm 1.45\)& 15.4\\
Power Loss (W)       & \(0.474 \pm 1.85\)& 37.4& \(0.542 \pm 1.57\)& 68.5\\
Total Power (W)      & \(7.96 \pm 8.76\)& 60.8& \(8.51 \pm 7.00\)& 90.4\\
Distance (m)& \(10.6 \pm 0.251\)& 11.3& \(9.61 \pm 0.518\)&12.0\\
Total Energy (J)& \(382 \pm 12.7\)& 407& \(409 \pm 14.5\)&439\\
MCOT                 & \(0.485 \pm 0.0256\)& 0.510& \(0.475 \pm 0.0209\)& 0.532\\ \bottomrule
\end{tabular}
\end{table}
 
\begin{table}[!ht]
\caption{Aggregate locomotion statistics across 100 traversals of 25 second segments at a fast walk of 0.4m/s, showing mean (\(\mu\)), standard deviation (\(\sigma\)), and maximum values.}
\label{tab:lunarleaper_traversal_fast}
\centering
\begin{tabular}{@{}lllll@{}}
\toprule
\textbf{Metric}            & \multicolumn{2}{l}{\textbf{Rigid terrain}} & \multicolumn{2}{l}{\textbf{Soft terrain}} \\ \midrule
                     & \(\mu\pm\sigma\)   & Max  & \(\mu\pm\sigma\)    & Max   \\ \midrule
Absolute Joint Torque (Nm) & \(0.519 \pm 0.743\)& 8.41& \(0.591 \pm  0.734\)& 6.62\\
Mechanical Power (W) & \(3.59 \pm 3.15\)& 19.6& \(3.03 \pm 2.37\)& 19.3\\
Power Loss (W)       & \(0.613 \pm 2.31\)& 49.3& \(0.653 \pm 1.86\)& 86.5\\
Total Power (W)      & \(10.9 \pm 11.8\)& 79.1& \(10.9 \pm 9.0\)& 107\\
Distance (m)& \(10.0 \pm 0.061\)& 10.2& \(10.1 \pm 0.18\)&10.5\\
Total Energy (J)& \(252 \pm 5.64\)& 260& \(250 \pm 11.7\)&278\\
MCOT                 & \(0.389 \pm 0.0246\)& 0.419& \(0.326\pm 0.0267\)& 0.386\\ \bottomrule
\end{tabular}
\end{table}

The comparison of rigid and soft terrain contact models for both speeds indicates that there is indeed a difference in both the qualitative and quantitative behaviours of the locomotion gait. The maximum joint torques are lower due to the lower impulse of soft contacts vs. the rigid contacts. Contrary to expectations, the mechanical power and MCOT is lower for the soft terrain.  However, the power losses are higher due to sustained foot contact forces in both the stance and swing stages of the gait. 

This is explained by the nature of the contact forces: a rigid contact model results in high impulses with rapid dissipation of kinetic energy, whereas a soft contact model has more gradual dissipation that varies linearly with depth. This, along with the dual effect of intrusion and extrusion forces for the soft model, causes the robot to lose more power due to the sustained effort required to push through the granular surface over time. Over time, this results in greater energy dissipation while covering the same target distance of 10 m. The maximum total energy and total power across all the simulations are also much higher, even though the mean and standard deviation are comparable. This indicates an increase in outlier cases where the quadruped is stuck in the regolith, possibly while trying to traverse a slope, requiring high power and energy consumption.
\section{Conclusion}
\label{sec:conclusion}
This report investigated the locomotion policy as trained by a Reinforcement Learning controller with a granular media terrain contact model. Through the implementation of the terradynamic model by Li et al. \cite{LiEtAl2013TerradynamicsLeggedLocomotion} in a simulation environment, the locomotion of a lunar explorer quadruped through lunar regolith was analysed. This ``soft'' granular terrain contact model was benchmarked against a rigid contact model (typically used in RL simulation) to determine the qualitative gait differences as well as quantitative differences in joint torques, power consumption, and energy.

The terradynamic model used was generalizable to different foot shapes and morphologies while also retaining the complexity of interactions of robot feet intruding into the ground at different angles and velocities. It allows for easy domain randomization of a single parameter to minimize the sim-to-real gap. In this implementation, it ran in a comparable computational time to the mesh point contact rigid contact solver. However, this model can result in occasional highly varying contact forces, which may cause flying or fast sliding behaviour. Further analysis of the intrusion and motion angle of the foot-surface contact for the foot morphology is required to establish numerical bounds for contact forces.

Benchmarking between a rigid contact and soft terrain contact model shows that traversal through granular substrates such as lunar regolith results in locomotion gait changes. The tracking ability is reduced due to slip issues during traversal. From a RL point of view, the space of feasible reward combinations is restricted due to model boundary issues; however, a common set of reward weights for both the rigid and soft contact models, which converge to a stable and energy efficient gait, can be found. Average and maximum torques increase, and although mechanical power is reduced, the resistance provided by the granular flow of the terrain results in greater power losses.

This exploration of different contact models demonstrates the necessity for granular terrain modelling in the mechanical and locomotion control co-design for legged robots in extra-terrestrial environments. An understanding of foot-terrain interactions for granular substrates is required for leg and foot morphology design and selection of motor drives, as well as robust and energy efficient locomotion control for such missions.

\section*{Acknowledgements}
The research leading to these results has received funding from the European Union's Horizon 2020 research and innovation programme under the Marie Sklodowska-Curie grant agreement no 101034319 and from the European Union — NextGenerationEU.

The author would like to thank Prof. Marco Hutter, Oliver Fischer, Joseph Church, Philip Arm, Adrian Fuhrer, Elena Krasnova, and Hendrik Kolvenbach at the Robotics Systems Lab for their guidance and support.

\bibliographystyle{ieeetr}
\bibliography{references}

\end{document}